%% file: preproc-ake.tex
\documentclass[11pt]{article}
\usepackage[utf8]{inputenc}
\usepackage{coling2016}
\usepackage{times}
\usepackage{url}
\usepackage{latexsym}
\usepackage{enumitem}
\usepackage{booktabs}
\usepackage{rotating}
\usepackage{graphics}
\usepackage{xcolor}

\colorlet{dr}{red!80!black}
\colorlet{dg}{green!60!black}

\title{How Document Pre-processing affects Keyphrase Extraction Performance}

\author{Florian Boudin \and Hugo Mougard \and Damien Cram\\
  LINA - UMR CNRS 6241, Université de Nantes, France \\
  {\tt firstname.lastname@univ-nantes.fr}
}

\date{}

\begin{document}

\maketitle

\begin{abstract}
The SemEval-2010 benchmark dataset has brought renewed attention to 
the task of automatic keyphrase extraction.
This dataset is made up of scientific articles that were automatically
converted from PDF format to plain text and thus require careful 
preprocessing so that irrevelant spans of text do not negatively affect
keyphrase extraction performance.
In previous work, a wide range of document preprocessing techniques were described but
their impact on the overall performance of keyphrase extraction models is still unexplored.
Here, we re-assess the performance of several keyphrase extraction models and measure
their robustness against increasingly sophisticated levels of document preprocessing.
\end{abstract}

\section{Introduction}

\blfootnote{
    %
    %
    %
    %
    \hspace{-0.65cm}  
    This work is licensed under a Creative Commons 
    Attribution 4.0 International Licence.
    Licence details:
    \url{http://creativecommons.org/licenses/by/4.0/}
    %
    %
}

Recent years have seen a surge of interest in automatic keyphrase
extraction, thanks to the availability of the SemEval-2010 benchmark 
dataset~\cite{kim-EtAl:2010:SemEval}.
This dataset is composed of documents (scientific articles) that were 
automatically converted from PDF format to plain text.
As a result, most documents contain irrelevant pieces of text (e.g.~muddled
sentences, tables, equations, footnotes) that require special handling, so as to 
not hinder the performance of keyphrase extraction systems.
In previous work, these are usually removed at the preprocessing step, but
using a variety of techniques ranging from simple
heuristics~\cite{wang-li:2010:SemEval,treeratpituk-EtAl:2010:SemEval,zervanou:2010:SemEval} to 
sophisticated document logical structure detection on richly-formatted documents
recovered from Google Scholar~\cite{nguyen-luong:2010:SemEval}.
Under such conditions, it may prove difficult to draw firm 
conclusions about which keyphrase extraction model performs best,
as the impact of preprocessing on overall performance cannot be properly quantified.

While previous work clearly states that efficient document preprocessing is a
prerequisite for the extraction of high quality keyphrases, there is,
to our best knowledge, no empirical evidence of how preprocessing
affects keyphrase extraction performance.
In this paper, we re-assess the performance of several state-of-the-art 
keyphrase extraction models at increasingly sophisticated levels of preprocessing.
Three incremental levels of document preprocessing are experimented with: raw text,
text cleaning through document logical structure detection, and removal of keyphrase
sparse sections of the document.
In doing so, we present the first consistent comparison of different 
keyphrase extraction models and study their robustness over noisy text.
More precisely, our contributions are:
\begin{itemize}

\item We show that performance variation across keyphrase extraction systems is, 
at least in part, a function of the (often unstated) effectiveness of document 
preprocessing.

\item We empirically show that supervised models are more resilient to noise, and
point out that the performance gap between baselines and top performing systems is
narrowing with the increase in preprocessing effort.

\item We compare the previously reported results of several keyphrase extraction 
models with that of our re-implementation, and observe that baseline performance
is underestimated because of the inconsistence in document preprocessing.


\item We release both a new version of the SemEval-2010 dataset\footnote{\url{https://github.com/boudinfl/semeval-2010-pre}} with
preprocessed documents and our implementation of the state-of-the-art keyphrase
extraction models\footnote{\url{https://github.com/boudinfl/pke}} using the \texttt{pke} toolkit~\cite{boudin:2016:ColingDemo} for use by the community.

\end{itemize}

%
%
%

\section{Dataset and Preprocessing}

The SemEval-2010 benchmark dataset~\cite{kim-EtAl:2010:SemEval} is composed of
244 scientific articles collected from the ACM Digital Library (conference
and workshop papers). 
The input papers ranged from 6 to 8 pages and were converted from PDF format
to plain text using an off-the-shelf tool\footnote{pdftotext,
\url{http://www.foolabs.com/xpdf/}}.
The only preprocessing applied is a systematic dehyphenation at line 
breaks\footnote{Valid hyphenated forms may have their hyphen removed by this process.}
and removal of author-assigned keyphrases.
Scientific articles were selected from four different research areas as defined
in the ACM classification, and were equally distributed into training (144
articles) and test (100 articles) sets.
Gold standard keyphrases are composed of both author-assigned keyphrases
collected from the original PDF files and reader-assigned keyphrases provided
by student annotators.

Long documents such as those in the SemEval-2010 benchmark dataset are 
notoriously difficult to handle due to the large number of keyphrase 
candidates (i.e.~phrases that are eligible to be 
keyphrases) that the systems have to cope with~\cite{hasan-ng:2014:P14-1}.
Furthermore, noisy textual content, whether due to format 
conversion errors or to unusable elements (e.g.~equations), yield many 
spurious keyphrase candidates that negatively affect keyphrase extraction 
performance.
This is particularly true for systems that make use of core NLP tools
to select candidates, that in turn exhibit poor performance on degraded text.
Filtering out irrelevant text is therefore needed for addressing these issues.

%
%
%
%


In this study, we concentrate our effort on re-assessing keyphrase
extraction performance on three increasingly sophisticated levels of 
document preprocessing described below.
%

\begin{enumerate}[wide=0.8em,leftmargin=0.8em,label=\textbf{Level \arabic*}:]

\item We process each input file with the Stanford CoreNLP suite~\cite{manning-EtAl:2014:P14-5}\footnote{Use use Stanford CoreNLP v3.6.0.}. We use the default parameters and perform tokenization, sentence splitting and Part-Of-Speech (POS) tagging.

\item Similarly to~\cite{nguyen-luong:2010:SemEval,lopez-romary:2010:SemEval}, we retrieve\footnote{To ensure consistency, articles were manually collected.} the original PDF files from the ACM Digital Library. We then extract the enriched\footnote{Additional information such as font format or spacial layout is also extracted.} textual content of the PDF files using an Optical Character Recognition (OCR) system\footnote{We use Omnipage v18, \url{http://www.nuance.com/omnipage}}, and perform document logical structure detection using ParsCit~\cite{Kan:2010:LSR:2436646.2436647}\footnote{We use ParsCit v110505.}.
%
%
We use the detected logical structure to remove author-assigned keyphrases and select only relevant elements: title, headers, abstract, introduction, related work, body text\footnote{We further filter out tables, figures, captions, equations, notes, copyright and references.} and conclusion.
We finally apply a systematic dehyphenation at line breaks and run the Stanford CoreNLP suite.

\item As pointed out by~\cite{treeratpituk-EtAl:2010:SemEval,nguyen-luong:2010:SemEval,wang-li:2010:SemEval,eichler-neumann:2010:SemEval,elbeltagy-rafea:2010:SemEval}, considering only the keyphrase dense parts of the scientific articles allows to improve keyphrase extraction performance.
Accordingly we follow previous work and further abridge the input text from level 2 preprocessed documents to the following: title, headers, abstract, introduction, related work, background and conclusion.
Here, the idea is to achieve the best compromise between search space (number of candidates) and maximum performance (recall).

\end{enumerate}

Table~\ref{tab:stats_on_train} shows the average number of sentences and words along with the maximum possible recall for each level of preprocessing.
The maximum recall is obtained by computing the fraction of the reference keyphrases that occur in the documents.
We observe that the level 2 preprocessing succeeds in eliminating irrelevant text by significantly reducing the number of words (-19\%) while maintaining a high maximum recall (-2\%).
Level 3 preprocessing drastically reduce the number of words to less than a quarter of the original amount while interestingly still preserving high recall.

\input{tables/stats.tex}

\section{Keyphrase Extraction Models}

We re-implemented five keyphrase extraction models~: the first two are commonly used as baselines, the third is a resource-lean unsupervised graph-based ranking approach, and the last two were among the top performing systems in the SemEval-2010 keyphrase extraction task~\cite{kim-EtAl:2010:SemEval}.
We note that two of the systems are supervised and rely on the training set to build their classification models.
Document frequency counts are also computed on the training set\footnote{For more reliable estimates, we rely on level 2 counts when experimenting with level 3.}.
Stemming\footnote{We use the Porter stemmer in nltk.} is applied to allow more robust matching.
The different keyphrase extraction models are briefly described below:

\begin{enumerate}[wide=0.8em,leftmargin=0.8em]

\item[\textbf{TF$\times$IDF}:] we re-implemented the TF$\times$IDF $n$-gram based baseline computed by the task organizers.
We use 1,~2,~3-grams as keyphrase candidates and filter out those shorter than 3 characters, containing words made of only punctuation marks or one character long\footnote{This filtering process is also applied to the other models.}.

\item [\textbf{Kea}~\cite{Witten:1999:KPA:313238.313437}:] keyphrase candidates are 1,~2,~3-grams that do not begin or end with a stopword\footnote{We use the stoplist in nltk, \url{http://www.nltk.org}}.
Keyphrases are selected using a na\"{i}ve bayes classifier\footnote{We use the Multinomial Naive Bayes 
classifier from scikit-learn with default parameters, \url{http://scikit-learn.org}} with two features: TF$\times$IDF and the relative position of first occurrence.

\item [\textbf{TopicRank}~\cite{bougouin-boudin-daille:2013:IJCNLP}:] keyphrase candidates are the longest sequences of adjacent nouns and adjectives. Lexically similar candidates are clustered into topics and ranked using TextRank~\cite{mihalcea-tarau:2004:EMNLP}. Keyphrases are produced by extracting the first occurring candidate of the highest ranked topics.

\item [\textbf{KP-Miner}~\cite{elbeltagy-rafea:2010:SemEval}:] keyphrase candidates are sequences of words that do not contain punctuation marks or stopwords.
Candidates that appear less than three times or that first occur beyond a certain position are removed\footnote{To better see the impact of preprocessing, we do not consider the cutoff parameter in our experiments. The least allowable seen frequency parameter is set to 2 which is the optimal value found on the training set.}.
Candidates are then weighted using a modified TF$\times$IDF formula that account for document length.

\item [\textbf{WINGNUS}~\cite{nguyen-luong:2010:SemEval}:] keyphrase candidates are simplex nouns and noun phrases detected using a set of rules described in~\cite{kim-kan:2009:MWE09}.
Keyphrases are then selected using a na\"{i}ve bayes classifier with the optimal set of features found on the training set\footnote{The optimal set of features in~\cite{nguyen-luong:2010:SemEval} also include the term frequency of substrings, but we observed a significant drop in performance when this feature is included.}: TF$\times$IDF, relative position of first occurrence and candidate length in words.

\end{enumerate}

Each model uses a distinct keyphrase candidate selection method that provides a trade-off between the highest attainable recall and the size of set of candidates.
Table~\ref{tab:recalls} summarizes these numbers for each model.
%
%
Syntax-based selection heuristics, as used by TopicRank and WINGNUS, are better suited to prune candidates that are unlikely to be keyphrases.
As for KP-miner, removing infrequent candidates may seem rather blunt, but it turns out to be a simple yet effective pruning method when dealing with long documents.
For details on how candidate selection methods affect keyphrase extraction, please refer to~\cite{Wang:2014:PAU:2649459.2649475}.

\input{tables/recalls.tex}

Apart from TopicRank that groups similar candidates into topics, the other models do not have any redundancy control mechanism.
Yet, recent work has shown that up to 12\% of the overall error made by state-of-the-art keyphrase extraction systems were due to redundancy~\cite{hasan-ng:2014:P14-1,boudin:2015:Keyphrase}.
Therefore as a post-ranking step, we remove redundant keyphrases from the ranked lists generated by all models.
A keyphrase is considered redundant if it is included in another keyphrase that is ranked higher in the list.


\section{Experiments}

\subsection{Experimental settings}

We follow the evaluation procedure used in the SemEval-2010 competition and evaluate the performance of each model in terms of f-measure (F) at the top $N$ keyphrases\footnote{Scores are computed using the evaluation script provided by the SemEval-2010 organizers.}.
We use the set of combined author- and reader-assigned keyphrases as reference keyphrases.
Extracted and reference keyphrases are stemmed to reduce the number of mismatches.

\subsection{Results}

The performances of the keyphrase extraction models at each preprocessing level are shown in Table~\ref{tab:results}.
Overall, we observe a significant increase of performance for all models at levels 3, confirming that document preprocessing plays an important role in keyphrase extraction performance.
Also, the difference of f-score between the models, as measured by the standard deviation $\sigma_1$, gradually decreases with the increasing level of preprocessing.
This result strengthens the assumption made in this paper, that performance variation across models is partly a function of the effectiveness of document preprocessing.

\input{tables/results.tex}

Somewhat surprisingly, the ordering of the two best models reverses at level 3.
%
%
%
This showcases that rankings are heavily influenced by the preprocessing stage, despite the common lack of details and analysis it suffers from in explanatory papers.
We also remark that the top performing model, namely KP-Miner, is unsupervised which supports the findings of~\cite{hasan-ng:2014:P14-1} indicating that recent unsupervised approaches have rivaled their supervised counterparts in performance.

In an attempt to quantify the performance variation across preprocessing levels, we compute the standard deviation $\sigma_2$ for each model.
Here we see that unsupervised models are more sensitive to noisy input, as revealed by higher standard deviations.
We found two main reasons for this.
First, using multiple discriminative features to rank keyphrase candidates adds inherent robustness to the models.
Second, the supervision signal helps models to disregard noise.
%

In Table~\ref{tab:pyramids}, we compare the outputs of the five models by measuring the percentage of valid keyphrases that are retrieved by all models at once for each preprocessing level.
By these additional results, we aim to assess whether document preprocessing smoothes differences between models.
We observe that the overlap between the outputs of the different models increases along with the level of preprocessing.
This suggests that document preprocessing reduces the effect that the keyphrase extraction model in itself has on overall performance. 
In other words, the singularity of each model fades away gradually with increase in preprocessing effort.

\input{tables/pyramids.tex}

\subsection{Reproducibility}

Being able to reproduce experimental results is a central aspect of the scientific method.
While assessing the importance of the preprocessing stage for five approaches, we found that several results were not reproducible, as shown in Table~\ref{tab:repro}.

\input{tables/repro-results.tex}

We note that the trends for baselines and high ranking systems are opposite: compared to the published results, our reproduction of top systems under-performs and our reproduction of baselines over-performs.
We hypothesise that this stems from a difference in hyperparameter tuning, including the ones that the preprocessing stage makes implicit.
Competitors have strong incentives to correctly optimize hyperparameters, to achieve a high ranking and get more publicity for their work while competition organizers might have the opposite incentive: too strong a baseline might not be considered a baseline anymore.

We also observe that with this leveled preprocessing, the gap between baselines and top systems is much smaller, lessening again the importance of raw scores and rankings to interpret the shared task results and emphasizing the importance of understanding correctly the preprocessing stage.

\section{Additional experiments}

In the previous sections, we provided empirical evidence that document preprocessing weighs heavily on the outcome of keyphrase extraction models.
This raises the question of whether further improvement might be gained from a more aggressive preprocessing.
To answer this question, we take another step beyond content filtering and further abridge the input text from level 3 preprocessed documents using an unsupervised summarization technique.
More specifically, we keep the title and abstract intact, as they are the two most keyphrase dense parts within scientific articles~\cite{nguyen-luong:2010:SemEval}, and select only the most content bearing sentences from the remaining contents.
To do this, sentences are ordered using TextRank~\cite{mihalcea-tarau:2004:EMNLP} and the less informative ones, as determined by their TextRank scores normalized by their lengths in words, are filtered out.

Finding the optimal subset of sentences from already shortened documents is however no trivial task as maximum recall linearly decreases with the number of sentences discarded.
Here, we simply set the reduction ratio to 0.865 so that the average maximum recall on the training set does not lose more than 5\%.
Table~\ref{tab:lvl4-stats} shows the reduction in the average number of sentences and words compared to level 3 preprocessing.

\input{tables/lvl4-stats.tex}

The performances of the keyphrase extraction models at level 4 preprocessing are shown in Table~\ref{tab:lvl4-results}.
We note that two models, namely TopicRank and TF$\times$IDF, lose on performance.
These two models mainly rely on frequency counts to rank keyphrase candidates, which in turn become less reliable at level 4 because of the very short length of the documents.
Other models however have their f-scores once again increased, thus indicating that further improvement is possible from more reductive document preprocessing strategies.

\input{tables/lvl4-results.tex}



\section{Conclusion}

In this study, we re-assessed the performance of several keyphrase extraction models and showed that performance variation across models is partly a function of the effectiveness of the document preprocessing.
Our results also suggest that supervised keyphrase extraction models are more robust to noisy input.

Given our findings, we recommend that future works use a common preprocessing to evaluate the interest of keyphrase extraction approaches. For this reason we make the four levels of preprocessing used in this study available for the community.

\bibliographystyle{coling2016}
\bibliography{biblio}

\end{document}

%% file: tables/stats.tex




\begin{table}[!ht]
    \centering
    \begin{tabular}{l | rrrr}
        ~ & \textbf{Lvl 1} & \textbf{Lvl 2} & \textbf{Lvl 3} \\ 
        \midrule
        Avg. sentences & 399    & 347    & 101    \\ 
        Avg. words     & 9\,772 & 7\,874 & 1\,922  \\ 
        Max. recall    & 83.9\% & 81.8\% & 70.9\% \\ 
    \end{tabular}
    \caption{Statistics computed at the different levels of document preprocessing on the training set.}
    \label{tab:stats_on_train}
\end{table}

%% file: tables/recalls.tex
\begin{table}[!ht]
    \centering
    \begin{tabular}{l | rr @{\hskip 1.5em} rr @{\hskip 1.5em} rr}
        \textbf{Model} & \multicolumn{2}{c}{\textbf{Lvl 1}} & \multicolumn{2}{c}{\textbf{Lvl 2}}
          & \multicolumn{2}{c}{\textbf{Lvl 3}} \\
        \midrule
        TF$\times$IDF & 80.2\% & 7\,837 & 78.2\% & 6\,958 & 67.8\% & 2\,270 \\
        Kea           & 80.2\% & 3\,026 & 78.2\% & 2\,502 & 67.8\% & 912    \\
        TopicRank     & 70.9\% & 742    & 69.2\% & 627    & 57.8\% & 241    \\
        KP-Miner      & 64.0\% & 724    & 61.8\% & 599    & 48.7\% & 212    \\
        WINGNUS       & 75.2\% & 1\,355 & 73.0\% & 1\,007 & 63.0\% & 403    \\
    \end{tabular}%
    \caption{Maximum recall and average number of keyphrase candidates for each model.}
    \label{tab:recalls}
\end{table}

%% file: tables/results.tex
\newcommand*\rot{\rotatebox{90}}

\begin{table}[htb!]
    \centering
    \begin{tabular}{ll| @{\hskip 0.5cm}p{1.2cm}p{1.2cm}p{1.2cm} | c}
    ~ & \textbf{Model} & \textbf{Lvl 1} & \textbf{Lvl 2} & \textbf{Lvl 3} & $\sigma_2$ \\
    \cmidrule[0.05em]{2-6}
    ~ & TopicRank                    & 12.2 & 12.5 & 14.5$^{\alpha,\beta}$ & 1.25 \\
    ~ & TF$\times$IDF                & 16.0 & 16.4 & 19.3$^{\alpha,\beta}$ & 1.80 \\
    \rot{\rlap{~Unsup.}} & KP-Miner  & 20.2 & 19.8 & \textbf{22.5}$^{\alpha,\beta}$ & 1.46 \\
    \cmidrule{2-6}
     & Kea                           & 19.2 & 19.3 & 21.2$^{\alpha}$ & 1.13 \\
    \rot{\rlap{Sup.}} & WINGNUS      & \textbf{20.5} & \textbf{20.3} & 21.8$^{\beta}$ & 0.82 \\
    \cmidrule[0.05em]{2-6}
    ~ & \multicolumn{1}{r|@{\hskip 0.5cm}}{$\sigma_1$} & 3.51 & 3.26 & 3.22 \\ 
    \end{tabular}
    \caption{F-scores computed at the top 10 extracted keyphrases for the unsupervised (Unsup.) and supervised (Sup.) models at each preprocessing level. We also report the standard deviation across the five models for each level ($\sigma_1$) and the standard deviation across the three levels for each model ($\sigma_2$). $\alpha$ and $\beta$ indicate significance at the 0.05 level using Student's t-test against level 1 and level 2 respectively.}
    \label{tab:results}
\end{table}


%% file: tables/pyramids.tex
\begin{table}[ht!]
    \centering
    \begin{tabular}{l|rrrr}
         & \textbf{Lvl 1} & \textbf{Lvl 2} & \textbf{Lvl 3} \\ 
        \midrule
        \% valid keyphrases & 19.9\% & 23.1\% & 25.1\% \\ 
    \end{tabular}
    \caption{Percentage of valid keyphrases found by all five keyphrase extraction models at each preprocessing level.}
    \label{tab:pyramids}
\end{table}


%% file: tables/repro-results.tex
\begin{table}[ht!]
    \centering
    \begin{tabular}{l|rrrr}
        \textbf{Model} & \textbf{Ori.} & \textbf{Lvl 1} & \textbf{Lvl 2} & \textbf{Lvl 3} \\ 
        \midrule
        TopicRank     & 12.1 & \textcolor{dg}{$+$0.1} & \textcolor{dg}{$+$0.4} & \textcolor{dg}{$+$2.4} \\ 
        TF$\times$IDF & 14.4 & \textcolor{dg}{$+$1.6} & \textcolor{dg}{$+$2.0} & \textcolor{dg}{$+$4.9} \\ 
        KP-Miner      & 23.2 & \textcolor{dr}{$-$3.2} & \textcolor{dr}{$-$3.4} & \textcolor{dr}{$-$0.7} \\ 
        WINGNUS       & 24.7 & \textcolor{dr}{$-$4.2} & \textcolor{dr}{$-$4.4} & \textcolor{dr}{$-$2.8} \\ 
    \end{tabular}
    \caption{Difference in f-score between our re-implementation and the original scores reported in~\protect\cite{hasan-ng:2014:P14-1,bougouin-boudin-daille:2013:IJCNLP}.}
    \label{tab:repro}
\end{table}

%% file: tables/lvl4-stats.tex
\begin{table}[!ht]
    \centering
    \begin{tabular}{l | rr}
        ~ & \textbf{Lvl 4} & \textbf{$\Delta$ Lvl 3} \\
        \midrule
        Avg. sentences & 71     & $-$30.0\%   \\
        Avg. words     & 1\,470 & $-$23.5\% \\
        Max. recall    & 65.9\% & $-$7.0\%    \\
    \end{tabular}
    \caption{Statistics computed at level 4 of document preprocessing on the training set. We also report the relative difference with respect to level 3 preprocessing.}
    \label{tab:lvl4-stats}
\end{table}

%% file: tables/lvl4-results.tex
\begin{table}[ht!]
    \centering
    \begin{tabular}{ll| rr}
    ~ & \textbf{Model} & \textbf{Lvl 4} & \textbf{$\Delta$ Lvl 3} \\
    \cmidrule[0.05em]{2-4}
    ~ & TopicRank                    & 13.7 & \textcolor{dr}{$-$0.8} \\
    ~ & TF$\times$IDF                & 18.5 & \textcolor{dr}{$-$0.8} \\
    \rot{\rlap{~Unsup.}} & KP-Miner  & 23.2 & \textcolor{dg}{$+$0.7} \\
    \cmidrule{2-4}
    ~ & Kea                          & 21.7 & \textcolor{dg}{$+$0.5} \\
    \rot{\rlap{Sup.}} & WINGNUS      & 22.5 & \textcolor{dg}{$+$0.7} \\
    \end{tabular}
    \caption{F-scores computed at the top 10 extracted keyphrases for the unsupervised (Unsup.) and supervised (Sup.) models at level 4 preprocessing. We also report the difference in f-score with level 3 preprocessing.}
    \label{tab:lvl4-results}
\end{table}